\def\year{2019}\relax
\documentclass[letterpaper]{article} 
\usepackage{aaai19}  
\usepackage{times}  
\usepackage{helvet}  
\usepackage{courier}  
\usepackage{url}  
\usepackage{graphicx}  
\usepackage{multirow}
\usepackage{amsfonts}
\usepackage{booktabs}
\usepackage{graphicx}
\usepackage{amsmath}
\usepackage{verbatim}
\usepackage{enumitem}
\usepackage{graphicx}

\usepackage{array}

\newcolumntype{C}[1]{>{\centering\arraybackslash}m{#1}}

\newcolumntype{L}[1]{>{\raggedright\arraybackslash}m{#1}}

\newcommand{\ie} {\emph{i.e. }}

\begin{document}

%

\title{Talking Face Generation by Adversarially Disentangled \\ Audio-Visual Representation}

\author{Hang Zhou, 
Yu Liu, Ziwei Liu\thanks{Corresponding author.}, Ping Luo, Xiaogang Wang\\
The Chinese University of Hong Kong, Hong Kong, China\\
\{zhouhang@link, yuliu@ee, zwliu@ie, xgwang@ee\}.cuhk.edu.hk, pluo.lhi@gmail.com}


\maketitle
\begin{abstract}
Talking face generation aims to synthesize a sequence of face images that correspond to a clip of speech.
This is a challenging task because face appearance variation and semantics of speech are coupled together in the subtle movements of the talking face regions.
Existing works either construct specific face appearance model on specific subjects or model the transformation between lip motion and speech. In this work, we integrate both aspects and enable arbitrary-subject talking face generation by learning disentangled audio-visual representation. We find that the talking face sequence is actually a composition of both subject-related information and speech-related information. These two spaces are then explicitly disentangled through a novel \textit{associative-and-adversarial} training process. This disentangled representation has an advantage where both audio and video can serve as inputs for generation. Extensive experiments show that the proposed approach generates realistic talking face sequences on arbitrary subjects with much clearer lip motion patterns than previous work. We also demonstrate the learned audio-visual representation is extremely useful for the tasks of automatic lip reading and audio-video retrieval.

\end{abstract}
\setcounter{secnumdepth}{1}
\section{Introduction}

\begin{figure}[t!]
\centering
\includegraphics[width=1\linewidth]{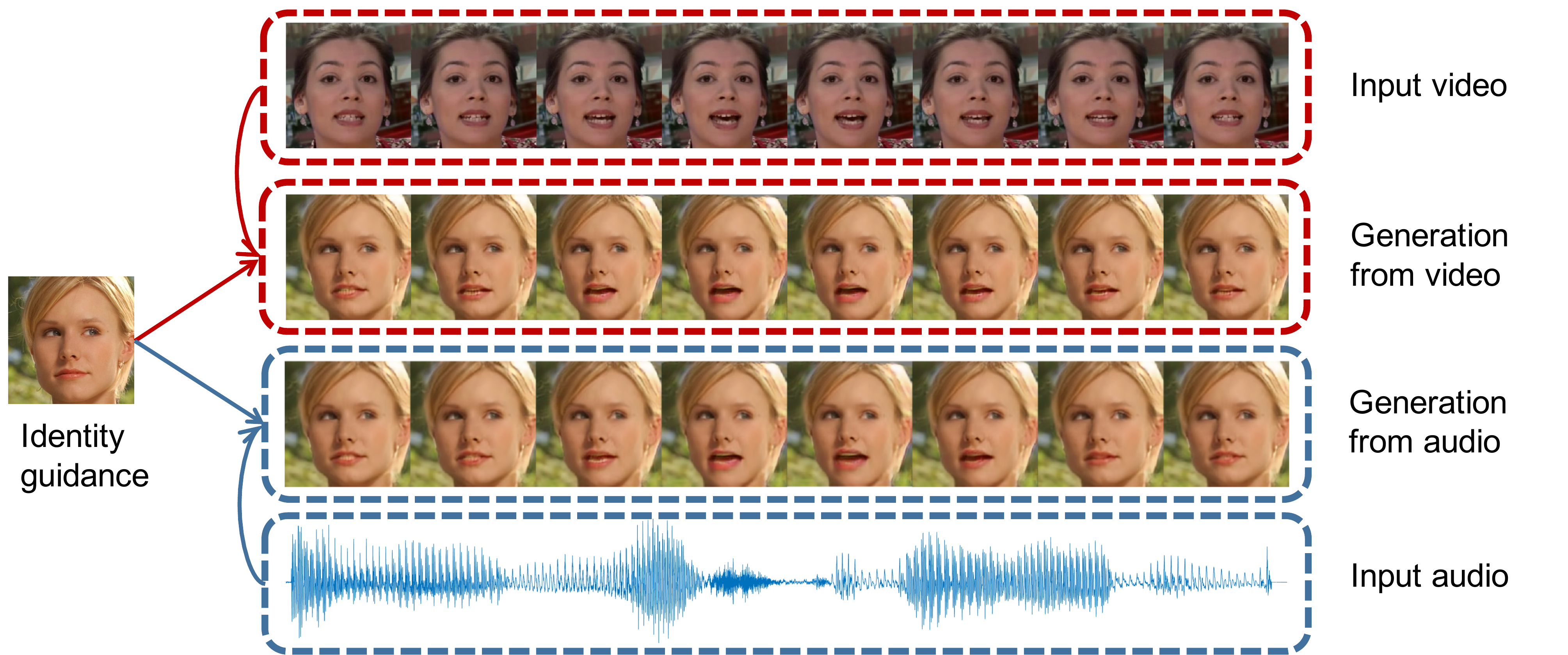}

\caption{Problem description. Given a single face image of a target person, this work aims to generate the talking video based on the given speech information that is represented by either a clip of video or an audio.}

\label{fig:problem}
\end{figure}

\noindent 
Understanding talking faces visually is of great importance to machine perception and communication. 
Humans can not only guess the semantic meaning of words by observing lip movement but also imagine the scenario when a specific subject talks (\ie face generation).
Recent advances have focused on automatic lip reading, which surpasses human-level performance in certain domains.
Here, we explore generating a video of arbitrary-subject speaking, which perfectly syncs with a specific speech where the speech information can be represented by either a clip of audio or video. We refer this problem as arbitrary-subject talking face generation, as shown in Fig.~\ref{fig:problem}.

However, generating identity-preserving talking faces that clearly conveys certain speech information is a challenging task, since the continuous deformation of the face region relates to both intrinsic subject traits~\cite{liu2015deep} and extrinsic speech vibrations.  
Previous efforts in this direction are mainly from computer graphics~\cite{xie2007realistic,wang2010synthesizing,fan2015photo,suwajanakorn2017synthesizing,thies2016face2face}. Researchers construct specific 3D face model for a chosen subject and the talking faces are animated by manipulating 3D meshes of the face model. 
However, these approaches strongly rely on the 3D face model and are hard to scale up to arbitrary identities.
More recent attempts~\cite{chung2017you} leverage the power of deep generative model and learn to generate talking faces from scratch.
Though the resulting models can be applied to an arbitrary subject, the generated face sequences are sometimes blurry and not temporally meaningful.
One important reason is that the subject-related and speech-related information are coupled together such that the talking faces are difficult to learn in a purely data-driven manner.

\begin{figure}[t!]
\centering
\includegraphics[width=1\linewidth]{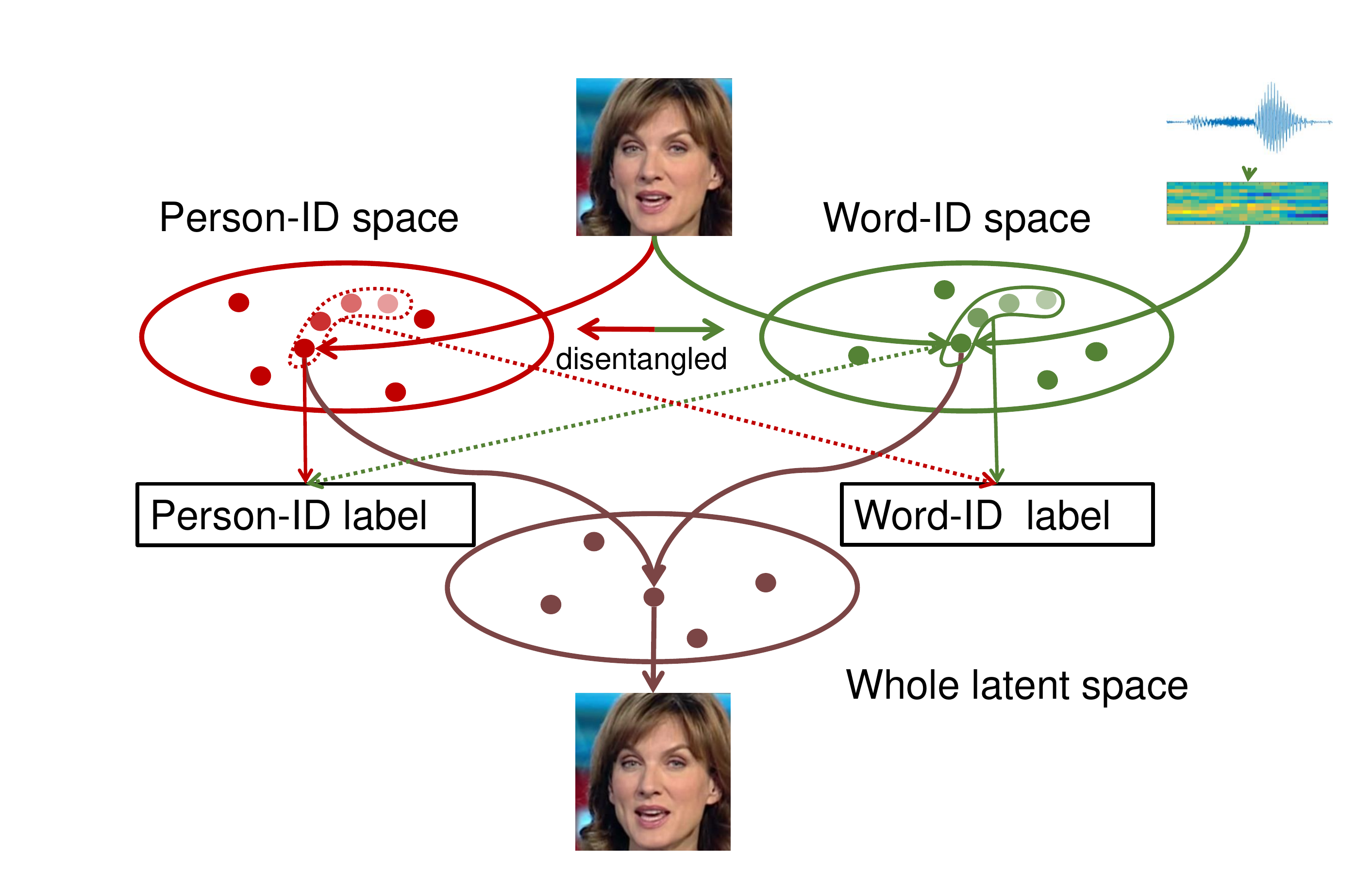}

\caption{We propose to guide the information flow by using labels to ensure the spaces contain discriminative semantic information dispelling from each other. With the assumption that Word-ID space is shared between visual and audio information, our model can reconstruct faces base on either video or audio.}
\label{fig:assumption}
\end{figure}

To address the aforementioned problems, we integrate the identity-related and speech-related information by learning disentangled audio-visual representation, as illustrated in Fig.~\ref{fig:assumption}.
We aim to disentangle a talking face sequence into two complementary representations, one containing identity information while the other containing speech information.
However, directly separating these two parts is not a trivial task because the variations of face deformation can be extremely large considering the diversity of potential subjects and speeches.

The key idea here is using audio-visual speech recognition~\cite{chung2016out,Chung17a} (\textit{i.e.} recognizing words from talking face sequence and audios, aka lip reading) as a probe task for \textit{associating} audio-visual representations, and then employing \textit{adversarial} learning to disentangle the subject-related and speech-related information inside them. 
Specifically, we first learn a joint audio-visual space where talking face sequence and its corresponding audio are embedded together.
It is achieved by enforcing the lip reading result obtained from talking faces aligns with the speech recognition result obtained from audio.
Next, we further utilize lip reading task to disentangle subject-related and speech-related information through adversarial learning~\cite{liu2018exploring}.
Notably, we enforce one of the representations extracted from talking faces to fool the lip reading system, in the sense that it only contains subject-related information, but not speech-related information.
Overall, with the aid of \textit{associative-and-adversarial} training, we can jointly embed audio-visual inputs and disentangle subject and speech-related information of talking faces.

The contributions of this work can be summarized as follows.
(1) A joint audio-visual representation is learned through audio-visual speech discrimination
 by associating several supervisions. Experiments show that the joint-embedding improves the baseline of lip reading result on LRW dataset~\cite{chung2016lip}.
(2) Thanks to the discriminative nature of our joint representation, we disentangle the person-identity and speech information through adversarial learning for better talking face generation.
(3) By unifying audio-visual speech recognition and audio-visual synchronizing, we achieve arbitrary-identity talking face generation from either video or audio speech as inputs in an end-to-end framework, which synthesizes high-quality and temporally-accurate talking faces.

\begin{figure*}[t!]
\centering
\includegraphics[width=1\linewidth]{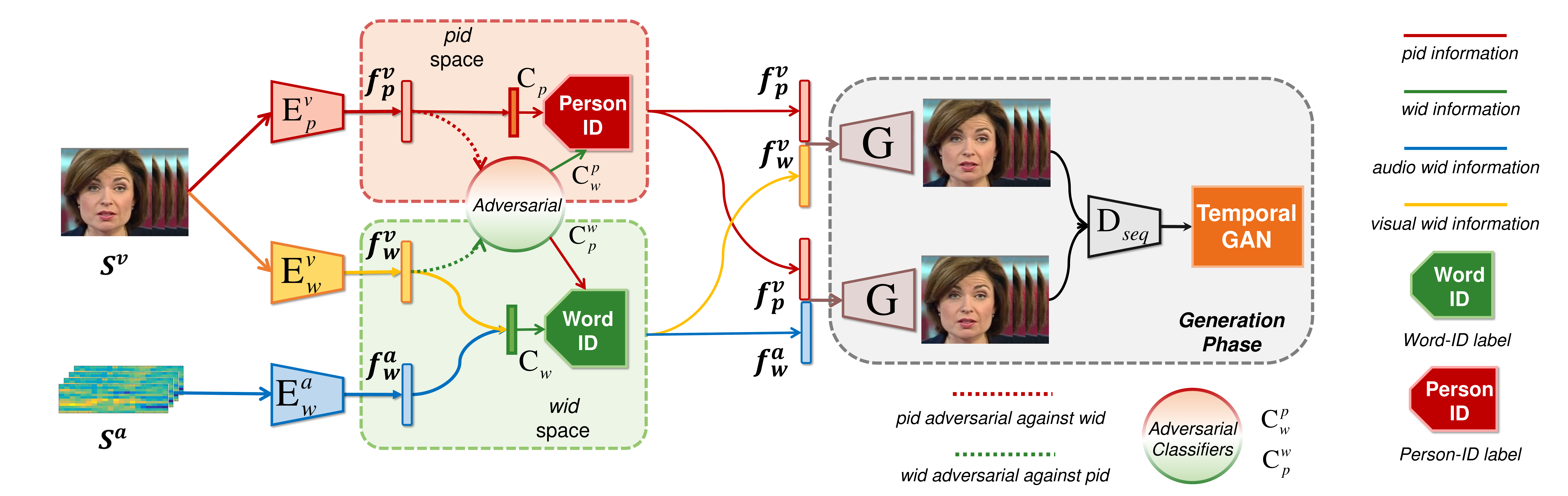}
\caption{Illustration of our framework. 
$\text{E}^v_{p}$ is the encoder that encodes Person-ID information from \textbf{visual} source to the $pid$ space, $\text{E}^v_{w}$ and $\text{E}^a_{w}$ are the Word-ID encoders that extract speech content information to $wid$ space from \textbf{video} and \textbf{audio}. Decoder $\text{G}$ takes any combination of features in $pid$ and $wid$ space to generate faces. $\text{D}_{seq}$ is a discriminator used for GAN loss. The adversarial training part contains two extra classifiers $\text{C}^w_p$ and $\text{C}^p_w$. The details of embedding the $wid$ space and adversarial training are shown in Fig~\ref{fig:partA} and~\ref{fig:aaai4}.}
\label{fig:framework}
\end{figure*}

\section{Related Work}
\label{RW}
\noindent\textbf{Generating Talking Faces.} 
The work of synthesizing lip motion from either audio~\cite{xie2007realistic,wang2010synthesizing,fan2015photo,fan2016deep,suwajanakorn2017synthesizing,chung2017you} or generating moving faces from videos~\cite{thies2016face2face,liu2017video,Wiles18a} has long been a task of concern in both the community of computer vision and graphics. However, most synthesis works from audio require a large amount of video footage of the target person for training, modeling, or sampling. They could not transfer the speech information to an arbitrary photo in the wild.

\cite{chung2017you} use a setting that is different from the traditional ones. They try to directly generate the whole face image with different lip motions in an image-to-image translation manner based on audios.
%
But their method base on data-driven training using an autoencoder, which leads to blurry results and lacks continuity. More recently, \cite{song2018talking} propose to use conditional RNN adversarial network, and~\cite{chen2018lip} propose to use correlation loss and three-stream GAN.
~\cite{Wiles18a} use flow to generate high precision arbitrary-identity talking face based on videos and claim to be able to produce videos based on audios, but with no results shown. However, as a common problem, without specific disentangling face and lip motion information, they all cannot generate high-quality results.
%

\noindent\textbf{Learning Audio-Visual Representation.}
The task of audio-visual speech recognition is a recognition problem uses either one or both video and audio as inputs.
Using visual information only for recognition is also referred to as \textit{Lip Reading}. A review of traditional methods for tackling this task has been made in~\cite{zhou2014review} thoroughly.
In recent years, this field develop quickly with the usage of convolutional neural networks (CNNs) and recurrent neural networks (RNNs) for end-to-end word-level~\cite{chung2016lip,stafylakis2017combining}, sentence-level~\cite{assael2016lipnet,chung2017lip}, and multi-view~\cite{Chung17a} lip reading.
In the meantime, the exploration of this topic has been greatly pushed forward by the build-up of large-scale word-level lip reading dataset~\cite{chung2016lip}, and the large sentence-level multi-view dataset~\cite{Chung17a}.

For the correspondence between human faces and audio clips, a number of works have been proposed to solve the problem of the audio-video synchronization between mouth motion and speech~\cite{mcallister1997lip,chung2016out}. Particularly, SyncNet~\cite{chung2016out,Chung17a} used two stream CNNs to sync audio \textit{mfcc} with 5 consecutive frames. In~\cite{Chung17a}, they further fixed the sync image feature as the pretraining for lip reading, but the two tasks are still separate from each other.
Recently, works from~\cite{nagrani2018seeing,nagrani2018learnable} also attempt to learn the association between a human face and voice for identity recognition instead of semantic level synchronization.

\section{Approach}
\label{PA}
We propose Disentangled Audio-Visual System (DAVS), an end-to-end trainable network for talking face generation by learning disentangled audio-visual representations, as shown in Fig.~\ref{fig:framework}. 

We leverage both talking video $S^v$ and its corresponding audio $S^a$ as training inputs. For learning the disentangled audio-visual representations between Person-ID space ($pid$) and the Word-ID space ($wid$), there are three encoder networks involved:
\begin{itemize}[leftmargin=*]
\item {\bf{Video}} to {\bf{Word-ID}} space encoder ($\text{E}^v_{w}$): $\text{E}^v_{w}$ learns to embed the video frame $s^v$ into a {\bf{visual}} representation $f^v_{w}$ which only contains speech-related information. It is achieved by learning a joint embedding space which \textit{associates} video and audio that correspond to the same word.   
\item {\bf{Audio}} to {\bf{Word-ID}} space encoder ($\text{E}^a_{w}$): $\text{E}^a_{w}$ learns to embed the speech $s^a$ into an {\bf{audio}} representation $f^a_{w}$, which resides in the shared space with $f^v_{w}$ as introduced above.
\item {\bf{Video}} to {\bf{Person-ID}} space encoder ($\text{E}^v_{p}$): $\text{E}^v_{p}$ learns to embed the video frame $s^v$ into a representation $f^v_{p}$ which only contains subject-related information. It is achieved by the \textit{adversarial} training process, forcing our target representation $f^v_{p}$ to fool the speech recognition system.
\end{itemize}

The whole idea of our pipeline is to first learn the discriminative audio-visual joint space $wid$, then disentangle it from the $pid$ space. Finally to combine features from the two spaces to get generation results.
Specifically, for learning the $wid$ space, we employ three supervisions: the supervision of Word-ID labels with shared classifier $\text{C}_w$ for associating audio and visual signals with semantic meanings; contrastive loss $\mathcal{L}_C$ for pulling paired video and audio samples closer; and an adversarial training supervision on audio and video features to make them indistinguishable. As for the $pid$ space, Person-ID labels from extra labeled face data are used.
For disentangling $wid$ and $pid$ spaces, adversarial training is employed.
As for generation, we introduce $L_{1}$-norm reconstruction loss $\mathcal{L}_{L_1}$ and temporal GAN loss $\mathcal {L}_{GAN}$ for sharpness and continuity.

\setcounter{secnumdepth}{2}
\subsection{Learning Joint Audio-Visual Representation}
\label{2.1}

\begin{figure}[t!]
\centering
\includegraphics[width=1\linewidth]{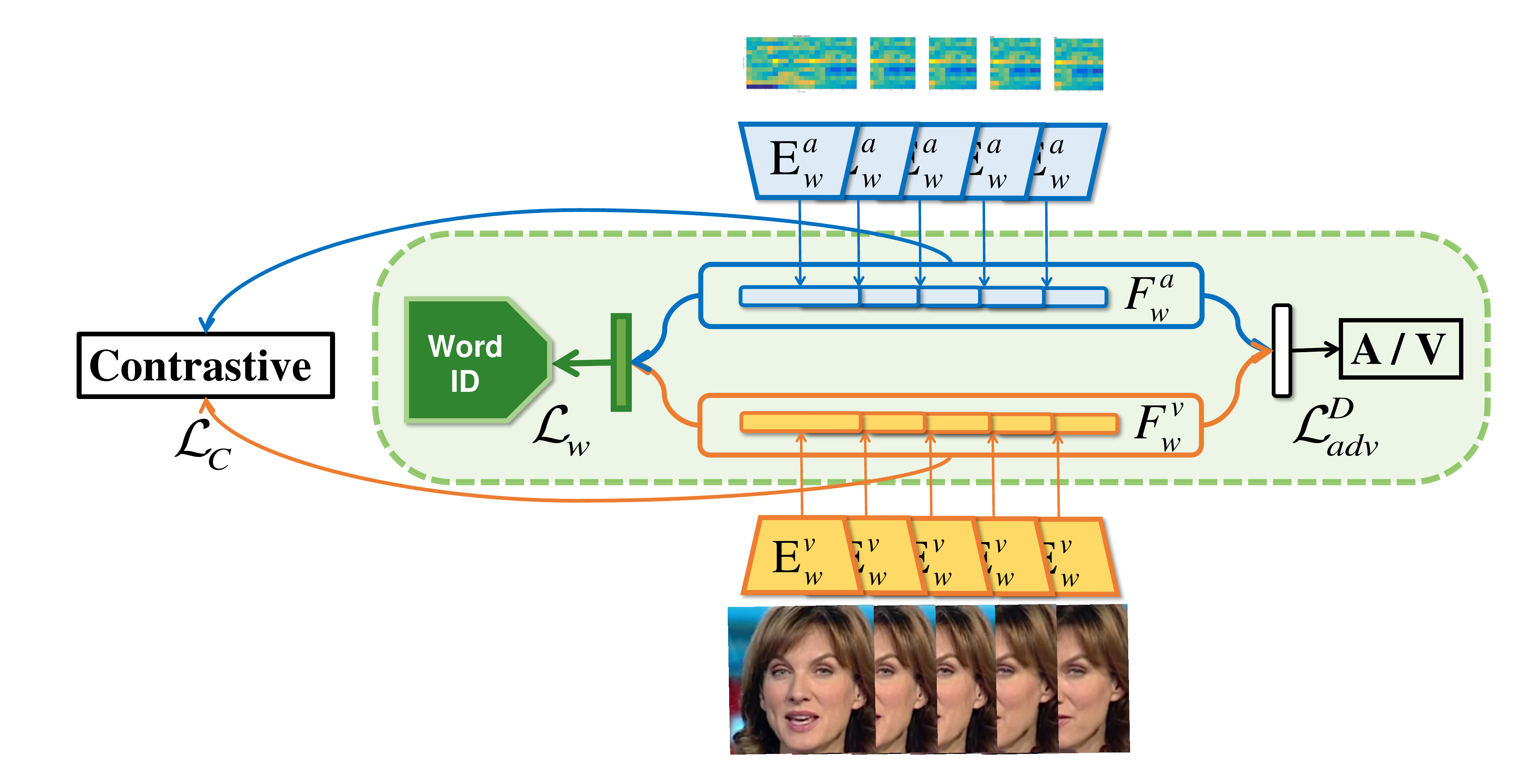}
\caption{Illustration of embedding the audio-visual shared space $wid$. The encoded features $F^v_{w} = [{f^v_{w}}_{(1)}, \cdots, {f^v_{w}}_{(n)}]$ and $F^a_{w} = [{f^a_{w}}_{(1)}, \cdots, {f^a_{w}}_{(n)}]$ are constrained by contrastive loss $\mathcal{L}_c$, classification loss $\mathcal{L}_w$ and domain adversarial training $\mathcal{L}^D_{adv}$.}
\label{fig:partA}
\end{figure}

We learn a joint audio-visual space that associates representations from both sources.
We constrain the extracted audio representation to be close to its corresponding visual representation, forcing the embedded features to share a same distribution and restricting $f^a_{w} \simeq f^v_{w} $, so that $G(f^v_{p}, f^v_{w}) \simeq G(f^v_{p}, f^a_{w})$ can be achieved. While requiring information of person facial identity flows from the $pid$ space, the other space of $wid$ would have to be person-ID invariant. The task of audio-visual speech recognition 
benefits us in achieving the shared latent space assumption and creating a discriminative space through mapping videos and audios to word labels. The implementation of learning the space is shown in Fig~\ref{fig:partA} (a). Then with the discriminative embedding, we can take the advantage of adversarial training for thoroughly information disentangling as described in Sec.~\ref{2.2}

\noindent\textbf{Sharing Classifier.}  
%
After the embedded features are extracted from the $wid$ encoders $\text{E}^a_{w}$, $\text{E}^v_{w}$ to get $F^v_{w} = [{f^v_{w}}_{(1)}, \cdots, {f^v_{w}}_{(n)}]$ and $F^a_{w} = [{f^a_{w}}_{(1)}, \cdots, {f^a_{w}}_{(n)}]$, normally they would be fed into different classifiers for visual and audio speech recognition. Here we share the classifier for both the modalities to enforce them to share their distributions. As a classifier's weight ${\bf{w}}_j$ tend to fall into the center of the clustering of the features belonging to the $j$'th class, through sharing the weights, the features between both modalities are pulled towards the centroid of the class~\cite{liu2018centroid}. The supervision is denoted as $\mathcal{L}_w$.

\noindent\textbf{Contrastive Loss.}  
As the problem of mapping audio and visual together is very similar to feature mapping~\cite{chopra2005learning}, retrieval and particularly the same as lip sync~\cite{chung2016out}, we adopted the contrastive loss which aims at bringing closer paired data while dispelling unpaired as a baseline. During training, for a batch of $N$ audio-video samples, the $m$th and $n$th sample are drawn with labels $l_{m = n} = 1$ while the others $l_{m \neq n} = 0$. The distance metric used to measure the distance between ${F^a_{w}}_{(m)}$ and ${F^v_{w}}_{(n)}$ here is the euclidean norm $d_{mn} = \|{F^v_{w}}_{(m)} - {F^a_{w}}_{(n)}\|_2$. The objective can be written as:
\begin{align}
    \label{eq1}
    \mathcal{L}_{C} = \sum_{n=1, m=1}^{N, N}{(l_{mn}d_{mn} + (1- l_{mn})\max(1 - d_{mn}, 0))} 
\end{align}
During our implementation, all features $F^v_{w}, F^a_{w}$ used in this loss are normalized first.

\noindent\textbf{Domain Adversarial Training.}  
To further push the face and audio features to be in the same distribution, we apply a domain adversarial training. An extra two-class domain classifier is appended for distinguishing the source of the feature. The audio and face encoders are then trained to prevent the classifier from success. This is mostly a simple version of the adversarial training described in section~\ref{2.2}. We refer to the objective of this method as $\mathcal{L}^D_{adv}$.

\subsection{Adversarial Training for Latent Space Disentangling}
\label{2.2}

In this section, we describe how we disentangle the subject-related and speech-related information in the joint embedding space using \textit{adversarial} training.

\begin{figure}[t!]
\centering
\includegraphics[width=1\linewidth]{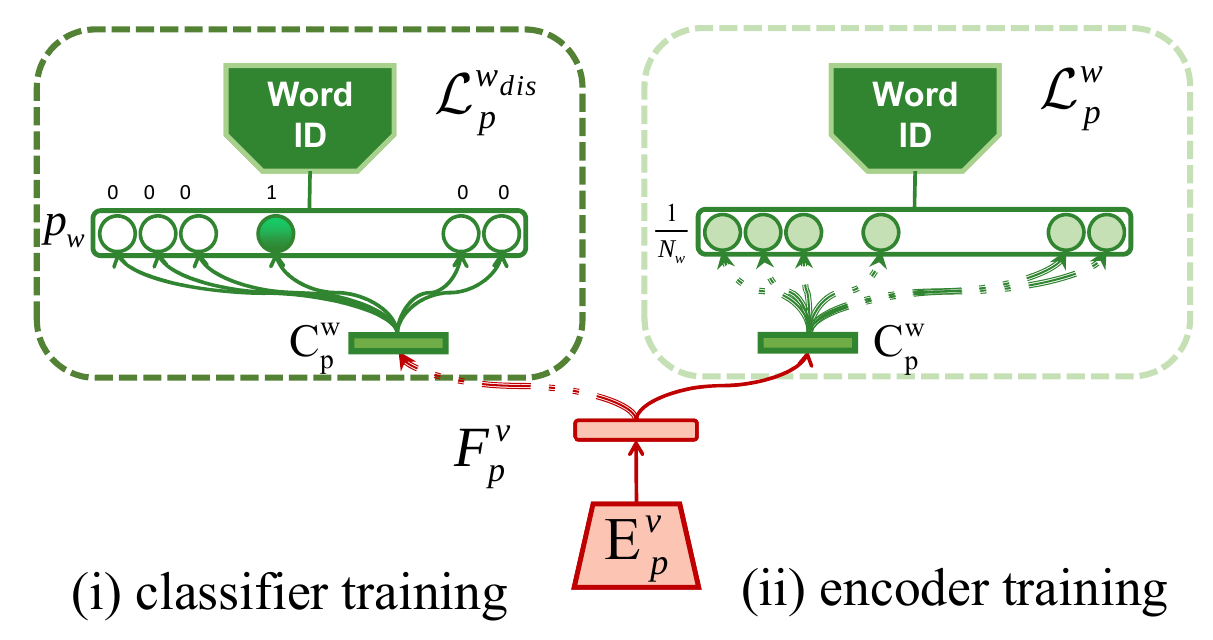}
\caption{Procedure of adversarial training for dispelling $wid$ information from $pid$ space. The training for  classifier $\text{C}^{w}_{p}$ is illustrated on the left and encoder $\text{E}^v_{p}$ on the right. The weights are updated on solid lines but not on the dashed lines.}
\label{fig:aaai4}
\end{figure}

Specifically, we would like the Person-ID feature $f^v_{p}$ to be free of Word-ID information. The discriminator could be formed to be a classifier $ \text{C}^w_p$ to map the collection of $F^v_{p} = [{f^v_{p}}_{(1)}, \cdots, {f^v_{p}}_{(n)}]$ to the $N_w$ Word-ID classes. The objective function for training the classifier is the same as softmax cross-entropy loss. However, the parameter updating is only performed on $\text{C}^w_p$, where ${p_w}^j$ is the one-hot label of the identity classes:
\begin{align}
	\label{eq2}
            \mathcal{L}^{w_{dis}}_p(\text{C}^{w}_{p}|\text{E}^v_{p}) = -\sum_{j=1}^{N_w}{p_{w}}^j\log(\text{softmax}(\text{C}^w_p(F^v_{p}))_j).
\end{align}
Then we update the encoder while fixing the classifier. 
The way to ensure that the features have lost all information about speech information is that it produces the same prediction for all classes after being sent into $\text{C}^{w}_{p} $. One way to form this limitation is to assign the probabilities of each word-label to be $\frac{1}{N_w}$ in softmax cross-entropy loss. The problem of this loss is that it would still backward gradient for updating parameters even if it reaches the minimum, so we propose to implement the loss using Euclidean distance:
\begin{align}
	\label{eq3}
            \mathcal{L}^{w}_p(\text{E}^v_{p}| {\text{C}}^{w}_{p}) = \sum_{j=1}^{N_w}\|\text{softmax}(\text{C}^w_p(F^v_{p}))_j - {\frac{1}{N_w}}\|^2_2.
\end{align}
The dual feature $f^v_{w}$ should also be free of $pid$ information accordingly, so the loss for encoding $pid$ information from each $f^v_{w}$ using classifier $\text{C}^p_w$ and loss for $wid$ encoder $\text{E}^v_{w} $ to dispel $pid$ information can be formed as follows:
\begin{align}
	\label{eq4}
            \mathcal{L}^{p_{dis}}_w(\text{C}^{p}_{w}|\text{E}^v_{w}) = -\sum_{j=1}^{N_p}{p_{p}}^j\log(\text{softmax}(\text{C}^p_w(f^v_{w}))_j),
\end{align}
\begin{align}
	\label{eq5}
            \mathcal{L}^{p}_w(\text{E}^v_{w}|\text{C}^{p}_w) = \sum_{j=1}^{N_p}\|{\text{softmax}}(\text{C}^{p}_w(f^v_{w}))_j - {\frac{1}{N_p}}\|^2_2.
\end{align}
$N_p$ is the number of person identities in the training set for embedding $pid$ space. We summarize the adversarial training procedure for classifier $\text{C}^w_p$ and encoder $\text{E}^v_p$ as Fig.~\ref{fig:aaai4}.

\begin{figure*}[t!]
\centering
\includegraphics[width=1\linewidth]{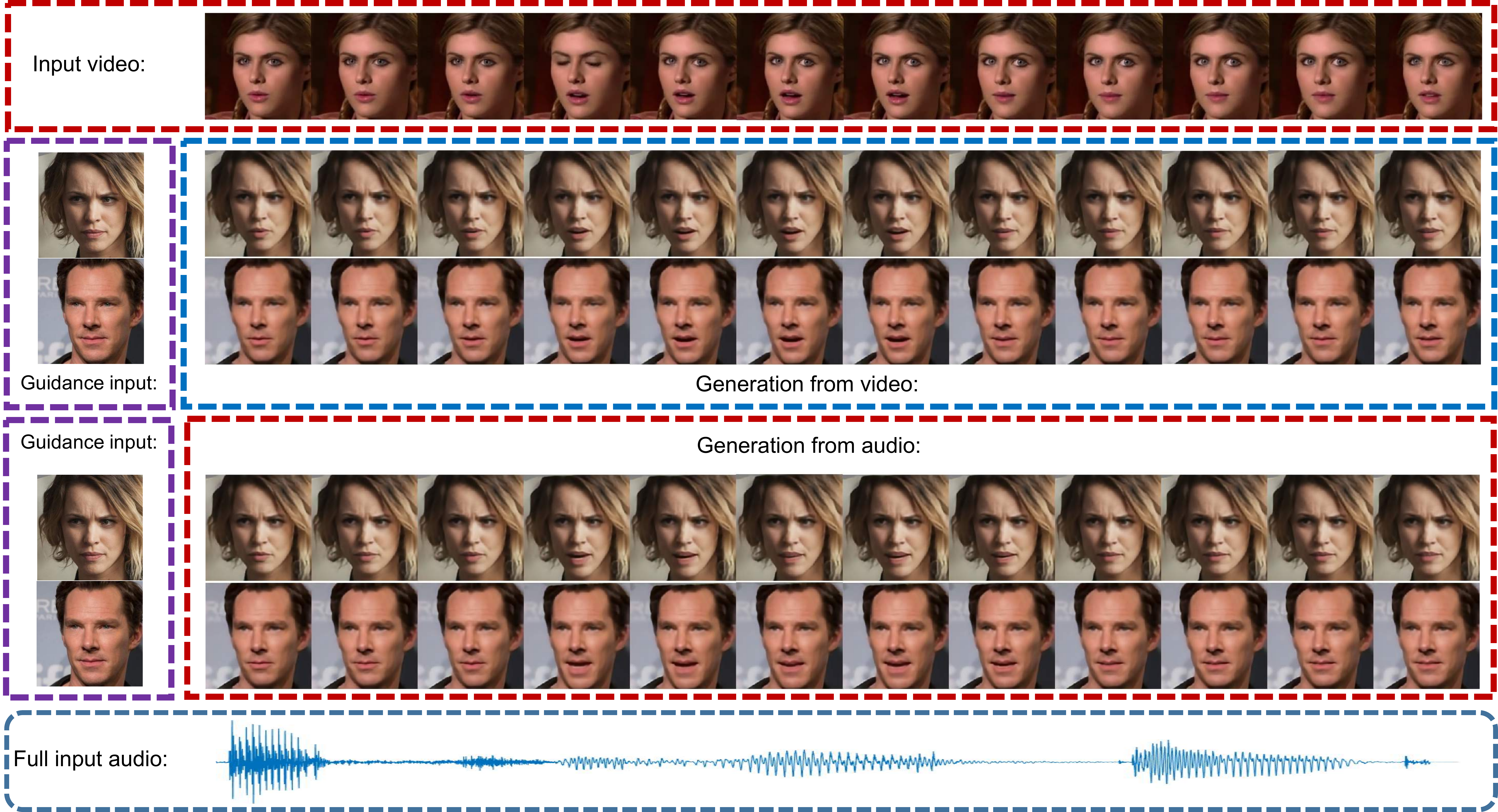}
\caption{Qualitative results. The guidance input image is on the left . The upper half is the generation from video and lower half is the generation from audio information.}
\label{fig:final2}
\end{figure*}

\subsection{Inference: Arbitrary-Subject Talking Face Generation}
In this section, we describe how we generate arbitrary-subject talking faces using the disentangled representations learned above. Combining $pid$ feature $f^v_{p}$ with either of the {\bf{video}} $wid$ feature $f^v_{w}$ or {\bf{audio}} $wid$ feature $f^a_{w}$, our system can generate a frame using the decoder G. The newly generated frame can be expressed as $\text{G}({f^v_{p}}, {f^v_{w}})$, $\text{G}({f^v_{p}}, {f^a_{w}})$.

Here we take synthesizing talking faces from audio $wid$ information as example. The generation results can be expressed as $\text{G}({f^v_{p}}_{(k)}, F^a_{w}) = \{\text{G}({f^v_{p}}_{(k)}, {f^a_{w}}_{(1)}), \cdots, \text{G}({f^v_{p}}_{(k)}, {f^a_{w}}_{(n)})\}$, where ${f^v_{p}}_{(k)}$ is the $pid$ feature of the random $k$th frame, which acts as identity guidance. Our overall loss function consists of a ${L}_1$ reconstruction loss and a temporal GAN loss, where a discriminator $\text{D}_{seq}$ takes the generated sequence $\text{G}({f^v_{p}}_{(k)}, F^a_{w})$ as input. These two terms can be formulated as follows:
\begin{align}
    \label{eq6}
{\mathcal {L}}_{L_1} = \| S^v - \text{G}({f^v_{p}}_{(k)}, F^a_{w}) \|_1,
\end{align}
\begin{align}
    \label{eq7}
    {\mathcal {L}}_{GAN} =\ &  \mathbb{E}_{S^v}[\log \text{D}_{seq}(S^v)]\ + \nonumber \\ 
    & \mathbb{E}_{F^v_{p}, F^a_{w}}[\log (1 - \text{D}_{seq}(\text{G}({f^v_{p}}_{(k)}, F^a_{w})]\
\end{align}
The overall reconstruction loss can be written as ${\mathcal {L}}_{Re}$, $\alpha$ is a hyper-parameter that leverages the two losses.
\begin{align}
    \label{eq8}
{\mathcal {L}}_{Re} = {\mathcal {L}}_{GAN} + \alpha {\mathcal {L}}_{L_1}.
\end{align}
The same procedure can be applied to generation from video information by substituting $F^a_{w}$ with $F^v_{w}$. As the reconstruction from audio and video can perform at the same time during training, we use ${\mathcal {L}}_{Re}$ to denote the overall reconstruction loss function.

\section{Experiments}

\noindent\textbf{Datasets.} 
Our model is trained and evaluated on the LRW dataset~\cite{chung2016lip}, which is currently the largest word-level lip reading dataset with $1$-of-$500$ diverse word labels. For each class, there are more than 800 training samples and 50 validation/test samples. Each sample is a one-second video with the target word spoken. Besides, the identity-preserving module of the network is trained on a subset of the MS-Celeb-1M dataset~\cite{guo2016ms}.
All the talking faces in the videos are detected and aligned using RSA algorithm~\cite{liu2017recurrent}, and then resized to $256 \times 256$. For the audio stream, we follow the implementation in~\cite{chung2016out} to extract the $mfcc$ features at the sampling rate of 100Hz. Then we match each image with a $mfcc$ audio input with the size of $12*20$.

\noindent\textbf{Network Architecture.} 
%
We adopted a modified VGG-M~\cite{Chatfield14} as the backbone for encoder $\text{E}^v_{p}$, and for encoder $\text{E}^v_{w}$, we modified a simple version of FAN~\cite{bulat2017far}. The encoder $\text{E}^a_{w}$ has a similar structure as that used in~\cite{chung2016out}.
Meanwhile, our decoder contains 10 convolution layers with 6 bilinear upsampling layers to obtain a full-resolution output image. All the latent representations are set to be 256-dimensional.

\noindent\textbf{Implementation Details.} We implemented DAVS using Pytorch. The batch size is set to be 18 with 1e-4 learning rate and trained on 6 Titan X GPUs. It takes about 4 epochs for the audio-visual speech recognition and person-identity recognition to converge and another 5 epochs for further tuning the generator. The whole training process takes about a week. Due to the alignment of the training set, the directly generated results may suffer from a scale changing problem, so we apply the subspace video stabilization~\cite{liu2011subspace} for smoothness.

\subsection{Results of Arbitrary-Subject Talking Face Generation}

At test time, the input identity guidance $s^v_p$ to $\text{E}^v_{p}$ is any person's face image and only one of the source for speech information $S^v_w$, $S^a_w$ is needed to generate a sequence of images. 

\noindent\textbf{Quantitative Results.} To verify the effectiveness of our GAN loss for improving image quality, we evaluate the PSNR and SSIM~\cite{wang2004image} score on the test set of LRW based on reconstruction. We compare the results with and without the GAN loss in Table~\ref{table:psnr}. We can see that both the scores are improved by changing $\mathcal{L}_{L_1}$ to $\mathcal{L}_{Re}$.

\begin{figure}[t!]
\centering
\includegraphics[width=1\linewidth]{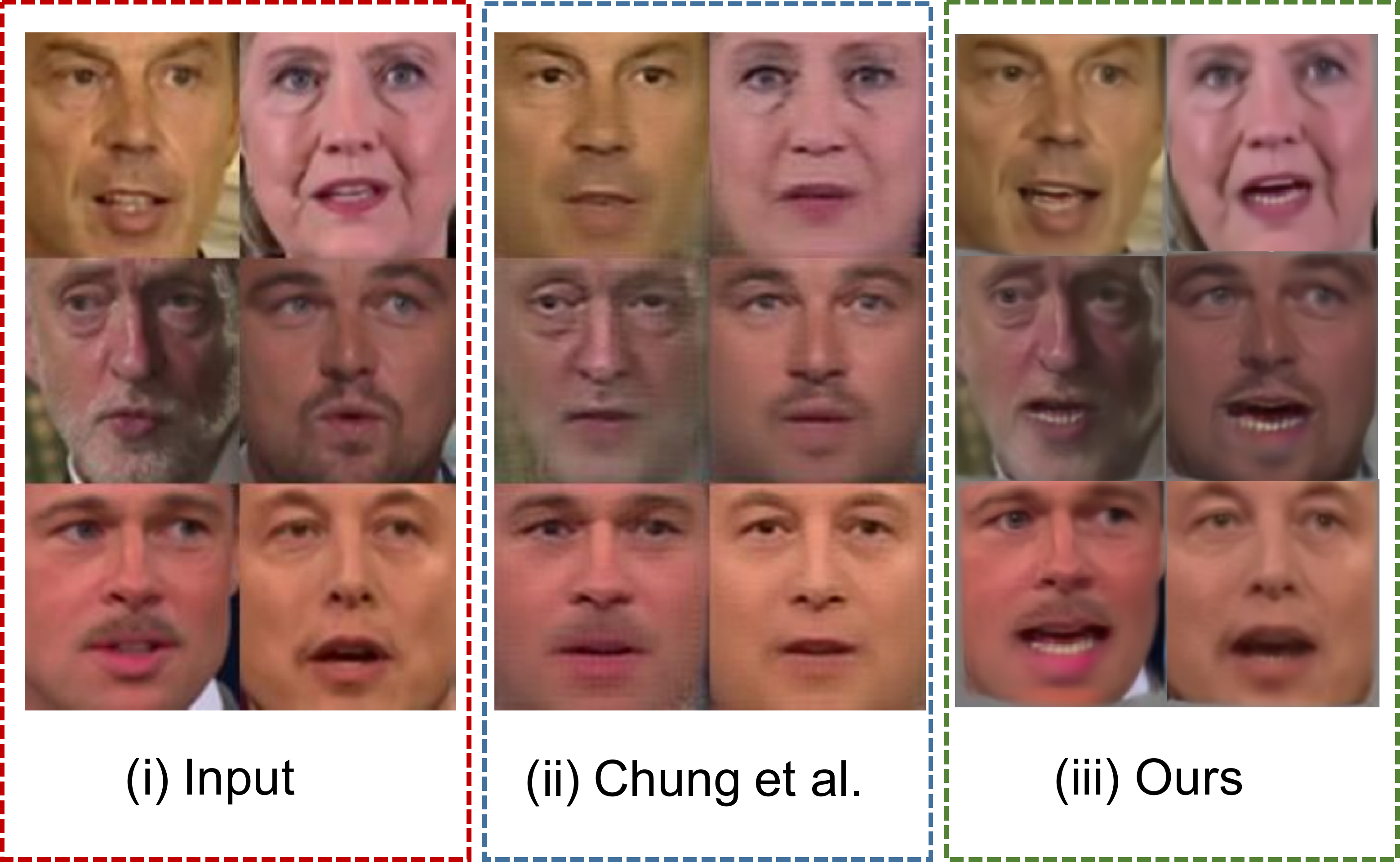}
\caption{Qualitative results comparing with Chung et al. The mouth shapes are arbitrary.}
\label{fig:aaai5}
\end{figure}

\noindent\textbf{Qualitative Results.} Video results are shown in {\textbf{supplementary materials}}. Here we show image results in Fig~\ref{fig:final2}. The input guidance photos are celebrities chosen randomly from the Internet. Our model is capable of generating talking faces based on both audios or videos. 
The focus of our work is to improve audio guided generation results by using joint audio-visual embedding, so we compare our work with \cite{chung2017you} at Fig~\ref{fig:aaai5}. It can be clearly seen that our results outperform theirs from both the perspective of identity preserving and image quality.

\setlength{\tabcolsep}{4pt}
\begin{table}[t] \normalsize 
\begin{center}

\caption{PSNR and SSIM scores for generation from audio and video $wid$ information with and without GAN loss.}
\label{table:psnr}
\begin{tabular}{ccc}
\toprule


~~~~~~~Approach  $\setminus$ Score & ~~~~~PSNR & ~~~~~~~~~~SSIM~~~~~~~~~~ \\
\noalign{\smallskip}

\midrule

~~~~~~~Audio (${\mathcal {L}}_{L_1}$) & ~~~~~25.4 & ~~~~~~~~~~0.859~~~~~~~~~~ \\
~~~~~~~Video (${\mathcal {L}}_{L_1}$) & ~~~~~25.7 & ~~~~~~~~~~0.865~~~~~~~~~~ \\
~~~~~~~Audio (${\mathcal {L}}_{Re}$) & ~~~~~\textbf{26.7} & ~~~~~~~~~~\textbf{0.883}~~~~~~~~~~ \\
~~~~~~~Video (${\mathcal {L}}_{Re}$) & ~~~~~\textbf{26.8} & ~~~~~~~~~~\textbf{0.884}~~~~~~~~~~ \\
 
\hline
\end{tabular}
\end{center}
\end{table}
\setlength{\tabcolsep}{1.4pt}


\setlength{\tabcolsep}{4pt}
\begin{table}[t] \small
\begin{center}

\caption{User study of our generation results and reproduced baseline. The results are averaged over person and time. }
\label{table:user}
\begin{tabular}{cccc}
\toprule

\noalign{\smallskip}

Method $\setminus$ Rate & Realistic & Lip-Audio Sync \\
\noalign{\smallskip}

\midrule
Reproduced Baseline  & 44.1\% & 58.0\% \\
Ours (Generation from Audio) & 51.5\% & 72.3\% \\
Ours (Generation from Video) & \textbf{87.8\%} & \textbf{88.4\%} \\
\hline
\end{tabular}
\end{center}
\end{table}
\setlength{\tabcolsep}{1.4pt}

\noindent\textbf{User Study.} 
We also conduct user study to investigate the visual quality of our generated results comparing with a fair reproduction of~\cite{chung2017you} with our network structure. They are evaluated \textit{w.r.t} two different criteria: whether participants could regard the generated talking faces as realistic (true or false), 
and how much percent of the time steps the generated talking faces temporally sync with the corresponding audio. 
We generate videos with the identity guidance to be 10 different celebrity photos. As for speech content information, we use clips from the test set of LRW dataset and selections from the Voxceleb dataset~\cite{Nagrani17}, which is not used for training. There are overall $10$ participants involved, and the results are average over persons and video time steps. The ground-truth is not included in the user study. Different subjects may behave different lip motion given the same audio clip and it is not desirable for the ground-truth to interfere with the participants' perception. 
When conducting the user study for lip sync evaluation, we asked the participants to only focus on whether the lip motion and given audio are temporally synchronized. Their ratings indicate that our generation results outperform the baseline by synchronizing rate and the extent of realistic, according to Table~\ref{table:user}. 

\setlength{\tabcolsep}{5.4pt}
\begin{table*}[t] \small

\begin{center}
\caption{Audio-Visual Speech Recognition and 1:25000 audio-video retrieval results with different supervisions. The first column is the supervisions, we use $\mathcal{L}_C$ to represent contrastive loss, SC for sharing classifier, $\mathcal{L}^D_{adv}$ for the adversarial training.}
\label{table:Co-embedding Results}
\begin{tabular}{lccccccccc}
\toprule
\noalign{\smallskip}
 & \multicolumn{3}{c}{Audio-Visual Speech Recognition} & \multicolumn{3}{c}{Video to Audio Retrieval} & \multicolumn{3}{c}{Audio to Video Retrieval}   \\
\noalign{\smallskip}

Approach &Visual acc.&Audio acc.&Combine acc.  & \textbf{$R@1$} & \textbf{$R@10$} & $Med \ R$ & \textbf{$R@1$} & \textbf{$R@10$} & $Med \ R$\\
\noalign{\smallskip}

\midrule
\cite{chung2016lip} & 61.1\% & - & - & - & - & - & - & - & - \\
Ours ($\mathcal{L}_C$)  & 61.8\% & 81.7 &90.8\%&29.3& 56.3& 6.0 & 29.8 &56.3 & 6.0\\
Ours ($\mathcal{L}_C$ + SC) & 65.6\% & 91.6\% & 94.9\% & 38.8  & 66.4 &  3.0 & 44.5 & 70.9 & 2.0\\
Ours ($\mathcal{L}_C$ + $\mathcal{L}^D_{adv}$) & 63.5\% & 88.1\% & 93.7\% & 39.3 & 67.9& 3.0 & 42.2 & 69.2 & 2.0\\
Ours ($\mathcal{L}_C$ + SC + $\mathcal{L}^D_{adv}$) & \textbf{67.5\%} & \textbf{91.8\%} &\textbf{95.2}\%& \textbf{64.2} & \textbf{84.7} & \textbf{1.0} & \textbf{67.7} & \textbf{85.8} & \textbf{1.0}\\

\hline
\end{tabular}
\vspace{-5pt}
\end{center}
\end{table*}

\setlength{\tabcolsep}{1.4pt}

\setlength{\tabcolsep}{5.4pt}
\begin{table*}[t] \footnotesize
\begin{center}
\caption{Ablation study on disentangle mechanism.}
\label{table:ablation}
\begin{tabular}{lcccccc}
\toprule
 & \multicolumn{3}{c}{Audio Generation to Source }& \multicolumn{3}{c}{Video Generation to Source}\\

Experiment &Retrieval~$R@1$ & Landmark $L_2$ &ID Squared $L_2$& Retrieval~$R@1$ & Landmark $L_2$&ID Squared $L_2$\\

\midrule
Direct Replication  & 2.5 & 4.27& - & 2.5 & 4.31 & - \\
Without Disentanglement & 53.8  & 3.94&  0.212 & 90.8 & 3.60 & 0.194\\
\textbf{With Disentanglement} & \textbf{60.5}  & \textbf{3.48} &\textbf{0.188} & \textbf{95.3} & \textbf{2.85} & \textbf{0.174}\\

\hline
\end{tabular}
\end{center}
\end{table*}
\setlength{\tabcolsep}{1.4pt}

\subsection{Effectiveness of Audio-Visual Representation}

In order to inspect the quality of our embedded audio-visual representation, we evaluate the discriminative power and the closeness of our co-embedded features.

\noindent\textbf{Word-level Audio-Visual Speech Recognition.} We report audio-visual speech recognition accuracy on the test set of LRW dataset. Containing the task of visual recognition (lip reading) and audio recognition (speech recognition).

Our model structure for lip reading is similar to the Multiple-Towers method which reaches the highest lip reading results in~\cite{chung2016lip}, so we consider it as a baseline. The difference is that the concatenation of features is performed at the spacial size of $1 \times 1$ in our setting. This would not be a reasonable choice for this task alone for the spatial information in images would be lost across time. However, as shown in Table~\ref{table:Co-embedding Results}, our results adding the contrastive loss alone outperforms the baseline. With the help of sharing classifier and domain adversarial training, the results improve a large margin.

\noindent\textbf{Audio-Video Retrieval.} To evaluate the closeness between the audio and face features, we borrow protocols used in the retrieval community. The retrieval experiments are conducted on the test set of LRW with 25000 samples, which means that given a test target video (audio), we try to find the closest audio (video) based on the distance of $wid$ features $F^v_{w}$, $F^a_{w}$ among all the test samples. 
Here we report the $R@1$, $R@10$ and $Med$ $R$ measurements which is the same as~\cite{faghri2017vse++}. 
As we can see in Table~\ref{table:Co-embedding Results}, with all supervisions, the highest results can be achieved.

\begin{figure}[t!]
\centering
\includegraphics[width=1\linewidth]{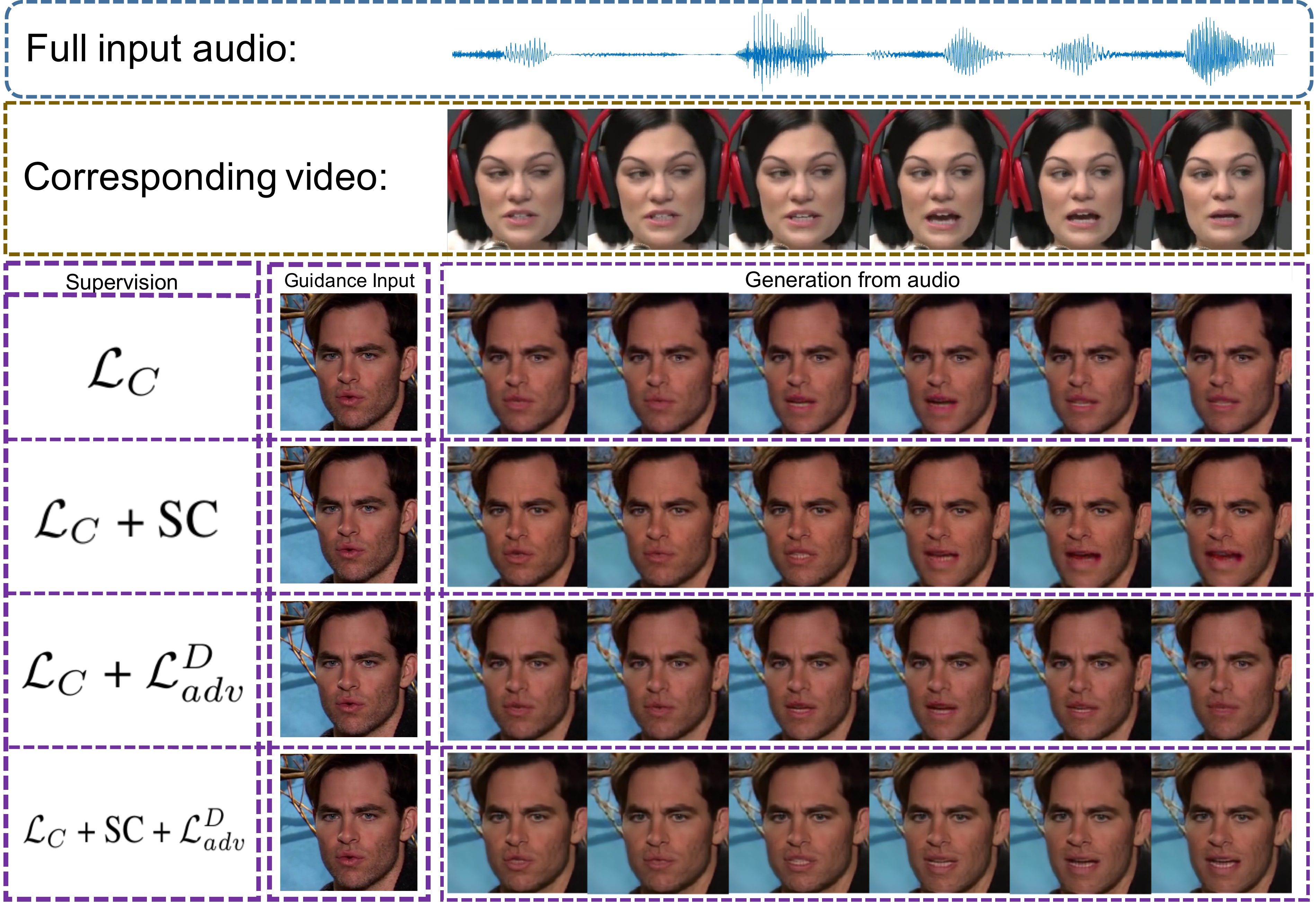}
\caption{Qualitative results for different types of supervisions. The left indicates different supervisions. All the generations are audio-based. }
\label{fig:aaai6}

\end{figure}

\noindent\textbf{Qualitative Results.}
Figure~\ref{fig:aaai6} shows the sequence generation quality from audio with different supervisions provided above. We can observe from the figure that given the same clip of audio, the duration of the mouth opening and to what extent it is opened is affected by different supervisions. Sharing the classifier apparently lengthens the time and strength of the mouth opening to make the image closer to the ground truth. Combining with the adversarial training makes the image quality improves. Note that it is not a one-to-one mapping between audio and lip motion; different subjects may behave different lip motion given the same audio clip so the final results may not perform the same as the ground truth.

\subsection{Identity-Speech Disentanglement} 

To validate our adversarial training is able to disentangle speech information from person-ID branch, we use person-ID encoder on every frame of a video and concatenate them to get $F^v_p = \{f^v_{p(1)}, ..., f^v_{p(n})\}$. 
Then we train an SVM to map training samples to their $wid$ labels and test the results, which implies that we attempt to find the $wid$ information left in the $pid$ encoder. The whole procedure is repeated before and after the feature disentanglement. Before the disentanglement, 27.8\% of the test set can be assigned to the right class, but only 9.7\% left after, indicating that considerable speech content information within the encoder $\text{E}^v_{p}$ is gone.

We then highlight the merits of adversarial disentanglement from two aspects, \textbf{identity preserving} and \textbf{lip sync quality}.
For identity preserving, we use OpenFace's squared L2 similarity score as an indicator and compare the identity distance between the generated faces and the original ones (lower indicates more similar).
For lip sync quality, we detect 20 landmarks using dlib library~\cite{dlib09} around the lips to characterize its deviation from ground truth, measured by the averaged L2-norm (lower is better).
Then we conduct retrieval experiments between all generated results and source videos based on extracted $F^v_{wid}$ features. Experiments are also conducted on a direct replication of every video clip, to prove that the retrieval results are affected by lip motion rather than appearance features. From Table~\ref{table:ablation}, we can observe that adversarial disentanglement indeed helps improves lip sync quality. 

\section{Conclusion}

In this paper, we propose a novel framework called Disentangled Audio-Visual System (DAVS), which generates high quality talking face videos using disentangled audio-visual representation. 
Specifically, we first learn a joint audio-visual embedding space $wid$ with discriminative speech information by leveraging the word-ID labels. 
Then we disentangled the $wid$ space from the person-ID $pid$ space through adversarial learning. 
Compared to prior works, DAVS has several appealing properties:
(1) A joint audio-visual representation is learned through audio-visual speech discrimination by associating several supervisions.
The disentangled audio-visual representation significantly improves lip reading performance;
(2) Audio-visual speech recognition and audio-visual synchronizing are unified in an end-to-end framework;
(3) Most importantly, arbitrary-subject talking face generation with high-quality and temporal accuracy can be achieved by our framework; both audio and video speech information can be employed as input guidance.

\section*{Acknowledgements}
We thank Yu Xiong for helpful discussions and his assistance with our video. This work is supported by SenseTime Group Limited, the General Research Fund sponsored by the Research Grants Council of Hong Kong 
and the Hong Kong Innovation and Technology Support Program (No.ITS/121/15FX).

\bibliographystyle{aaai}
\bibliography{egbib}
\end{document}